\documentclass[default,iicol,sn-aps]{sn-jnl}

\jyear{2022}%

\theoremstyle{thmstyleone}%
\theoremstyle{thmstyletwo}%

\theoremstyle{thmstylethree}%

\begin{document}

\title[Article Title]{Neural Generalized Ordinary Differential Equations with Layer-varying Parameters}


\author[1]{\fnm{Duo} \sur{Yu}}\email{duo.yu@austin.utexas.edu}

\author[2]{\fnm{Hongyu} \sur{Miao}}\email{hmiao@fsu.edu}

\author*[3]{\fnm{Hulin} \sur{Wu}}\email{hulin.wu@uth.tmc.edu}

\affil[1]{\orgdiv{Department of Population Health}, \orgname{The University of Texas at Austin}, \orgaddress{\street{1601 Trinity St.}, \city{Austin}, \postcode{78712}, \state{Texas}, \country{United States}}}

\affil[2]{\orgdiv{College of Nursing}, \orgname{Florida State University}, \orgaddress{\street{98 Varsity Way}, \city{Tallahassee}, \postcode{32306}, \state{Florida}, \country{United States}}}

\affil*[3]{\orgdiv{Department of Biostatistics and Data Science}, \orgname{The University of Texas Health Science Center at Houston}, \orgaddress{\street{1200 Pressler  Street}, \city{Houston}, \postcode{77030}, \state{Texas}, \country{United States}}}


\abstract{Deep residual networks (ResNets) have shown state-of-the-art performance in various real-world applications.  Recently, the ResNets model was reparameterized and interpreted as solutions to a continuous ordinary differential equation or Neural-ODE model. In this study, we propose a neural generalized ordinary differential equation (Neural-GODE) model with layer-varying parameters to further extend the Neural-ODE to approximate the discrete ResNets. Specifically, we use nonparametric B-spline functions to parameterize the  Neural-GODE so that the trade-off between the model complexity and computational efficiency can be easily balanced. It is demonstrated that ResNets and Neural-ODE models are special cases of the proposed Neural-GODE model. Based on two benchmark datasets, MNIST and CIFAR-10, we show that the layer-varying Neural-GODE is more flexible and general than the standard Neural-ODE. Furthermore, the Neural-GODE enjoys the computational and memory benefits while performing comparably to ResNets in prediction accuracy.}




\maketitle

\section{Introduction}\label{sec1}

Deep learning (or deep neural networks) has been successfully applied in a variety of real-world areas, including natural language processing \cite{graves2013speech,bahdanau2014neural, young2018recent}, computer vision \cite{krizhevsky2009learning, goodfellow2014generative,long2015fully}, speech recognition \cite{noda2015audio, deng2014ensemble, yu2016automatic}, et al.. One type of deep learning model, called deep residual networks (ResNets), has shown state-of-the-art performance in image recognition \cite{he2016deep, qiu2017learning, zhang2017residual}. By incorporating the shortcut connection, the ResNets improves learning performance with deeper and wider architectures \cite{bishop1995neural, ripley2007pattern}. It has been considered as a default practice of the convolutional neural networks (CNN) models and a powerful tool to deal with complex image recognition problems. For example, ResNets-152 achieves 19.38\% top-1 error on the ImageNet data with 152 layers \cite{he2016deep}; and ResNets-1001 reaches 4.92\% test error on the CIFAR-10 data  with 1000 layers \cite{he2016identity}.

A ResNets transforms the input and hidden layers iteratively to filter the information:
\begin{equation}
    \label{restrans}
    h_{t+1}=h_t + \sigma(h_t, \theta_t) \quad \text{for}\quad t = 0, 1, \dots, T-1
\end{equation}
where $h_0$  is the input data, $h_1,h_2, \dots, h_T$ are the hidden layers,  $\sigma$ denotes the activation function, $\theta_t$ represents the parameters that link $h_t$ and $h_{t+1}$. These iterative updates can be interpreted as an Euler discretization of a nonlinear ordinary differential equation (ODE) \cite{haber2017stable, li2017maximum}:
\begin{equation}
    \label{ode}
    h'(t) =\sigma(h(t),\theta (t))
\end{equation}
Based on this interpretation, many novel deep learning models and training algorithms have been proposed \cite{chang2018reversible, haber2017stable, lu2018beyond, li2017maximum, chen2018neural}. For example, Chen et al. proposed the Neural-ODE model that replaced the multiple residual blocks with one ODE system in the model architecture \cite{chen2018neural}. To train such a Neural-ODE model, the authors developed the publicly available software, torchdiffeq, which adopted various ODE solvers at the back-propagation step. Compared with the original ResNets, many advantages of the proposed Neural-ODE were demonstrated, such as memory efficiency, adaptive computation of solving ODEs, scalable and invertible normalizing flow constructions, and building continuous time-series models.

Conceptually, the Neural-ODE model combines the strengths of both parametric and nonparametric approaches for model construction and training. Parametric dynamical models, such as ordinary differential equations (ODEs), have long been studied in, e.g., mathematics, physics, and engineering; therefore, a rich amount of theories and application experiences have been accumulated \cite{lasalle1968stability, simmons2016differential, arnold2012geometrical, yu2016revisiting,yu2017effects, yu2021assessing}. In general, the ODE models use parameterized mathematical functions to describe the changes of crucial variables given a dynamic system, such as the logistic model for population growth \cite{yu2016revisiting}, prey-predator dynamics in ecological studies \cite{tang2015holling}, and Susceptible-Infectious-Recovered (SIR) model in infectious disease transmission \cite{yu2017effects}. Then the parameters of the ODE model are estimated based on observed data so that the proposed ODE can describe the dynamic behavior of a system. However, this parametric approach is limited by its assumption of the underlying mechanisms, which may partially capture the real dynamics. On the other hand, the nonparametric models, such as recurrent neural networks (RNNs) \cite{sherstinsky2020fundamentals}, multilayer perceptron (MLP), and deep residual networks (ResNets), have been successfully applied in a variety of areas due to their universal approximation property. By connecting the deep residual networks and ordinary differential equations, the Neural-ODE model serves as a promising model to better capture the real-world dynamics. 

Previously, we have proposed a generalized ODE (GODE) to model the dynamics of discrete data \cite{miao2014generalized}. Specifically, the GODE model can apply to a wide range of data types, including those that follow the exponential family distribution, and the associated link function is capable of accommodating latent time-varying variables governed by ODEs. Note that the latent time-varying variables in GODE can be considered as a hidden layer in neural networks. Especially, it is equivalent to the hidden layer of ODE block in the Neural-ODE \cite{chen2018neural}. In practice, the ordinary differential equations (ODEs) with time-varying coefficients \cite{chen2008efficient, xue2010sieve} could be used to describe the time-varying changes of model parameters, which results in more flexible dynamic systems. This type of dynamic model has been widely used in many biomedical applications \cite{chen2008efficient, xue2010sieve, liang2010estimation}. In this study, we coupled the time-varying ODE and GODE ideas to generalize the standard Neural-ODEs model proposed by Chen et al. \cite{chen2018neural} and propose a neural generalized ODE (Neural-GODE) model. The performance of the proposed Neural-GODE is evaluated and compared with existing deep residual networks (ResNet) \cite{he2016deep} and the standard Neural-ODEs model \cite{chen2018neural} using benchmark datasets, MNIST \cite{lecun1998gradient} and CIFAR-10 \cite{krizhevsky2009learning}.

\section{Method}\label{sec2}
Given a supervised learning problem with input $ \mathcal{X}$ and output label $ \mathcal{Y}$, the essential task is to find a mapping 
\begin{equation}
    \label{supmach}
    F: \mathcal{X} \to \mathcal{Y}
\end{equation}
where $\mathcal{X} \subset \mathbb{R}^p$, $\mathcal{Y} \subset \mathbb{R}$, such that $F(\mathbf{X}_i)$ can accurately predict $y_i$, and $(\mathbf{X}_i,y_i)$ is the $i$-th sample, $i=1, 2, \dots, n$. Usually, $F$ is approximated by penalized regression, algorithms, and networks, such as Lasso, support vector machine (SVM), and multilayer perceptron (MLP). Especially with the universal approximation property, multilayer feedforward networks, such as residual neural networks (ResNets), multilayer perceptron (MLP), convolutional neural networks (CNN), are widely applied in complex prediction tasks \cite{hornik1989multilayer}. Multilayer feedforward networks use the forward propagation technique that processes the inputs in a nonlinear and forward direction way to filter the information. 
For example, in a general residual neural networks (ResNets), the forward propagation of input $Z_0 \in R^{n\times p}$, with $T$ layers is given by
\begin{equation}
    \label{restrans1}
   Z_{(t+1)}=Z_t+h\sigma (Z_t K_t+b_t ) \quad \text{for} \quad t=0,1, \dots,T-1, 
\end{equation}
where $Z_0=X$, and $Z_1,Z_2,\dots,Z_T$ are the hidden layers, $t$ is the layer index, $\sigma$ is the activation function, and $h$ is the scaling factor, $K_t$ and $b_t$ are the constant weights of the $t$-th hidden layer. The iterative updates of hidden layers $Z_t$, equation (\ref{restrans1}),  can be interpreted as a discretized nonlinear ordinary differential equation (ODE):
\begin{equation}
    \label{ode2}
    \dot{Z}(t)=\sigma{Z(t)\beta(t)+b(t)}
\end{equation}
where $Z(t)$ is a time-varying variable with initial $Z(0)=X$; $\beta(t)$ and $b(t)$  are time-varying parameters.
Assume the outcome $Y$ follows an exponential family distribution with mean of $E(Y)= \mu $ and link function involving a hidden variable $Z_T$:
\begin{equation}
    \label{link}
    \eta = g(\mu)=g^{*} (Z_T,\theta),
\end{equation}
where $Z_T \doteq Z(T)$ is the solution to the ODE in equation (\ref{ode2}). The optimization problem with respect to $\beta(t), b(t), W, b$ can be written as 
\begin{align}
    &\text{min}\frac{1}{n}L(\beta(t), b(t), W, b\mid X, Y) +\lambda P(\beta(t), b(t), W, b) \nonumber \\
    & \text{s.t.} \quad \dot{Z}(t)=\sigma{Z(t)\beta(t)+b(t)}, \quad Z(0)= X,  \nonumber \\
    &  \eta = g(\mu)=g^{*} (Z_T,\theta). \label{opt}
\end{align}
 Here, $L(\beta(t),b(t),W,b\mid X,Y) $ is the likelihood function, and $P(\beta(t), b(t), W, b)$ is the penalty function, $\lambda$ is the penalty parameter. Adopting the idea of the time-varying ODE model in Xue et al. \cite{xue2010sieve}, we can approximate the time-varying parameters using B-splines, i.e., 
 \begin{align}
     \beta(t) &= B_1 (t)\xi, \nonumber \\
b(t)&= B_2 (t)\zeta, \label{timevary}
 \end{align}
where $B_1 (t)$ and $B_2 (t)$ are B-spline basis functions and $(\xi, \zeta)$ are constant parameters. A brief review of the B-spline will be shown in the next subsection.

\subsection{B-spline function}\label{subsec21}
The spline functions have been widely used in nonparametric regression and varying-coefficient models in statistical research \cite{perperoglou2019review}. In particular, the splines are generally used for modeling the smooth functions of the interested variables, such as nonlinear effects of covariates, time-dependent effects in regression models, and time-series data modeling.

B-splines are a more general type of curve than B$\acute{e}$zier curve \cite{bartels1995introduction}. A B-spline with degree k and n control points can be written as 
\begin{equation}
    \label{bspline}
    \beta(t)= \sum_{i=0}^{n-1}B_{i,k} (t)\xi_i,
    \end{equation}
where $B_{i,k} (t)$ is the degree $k$ basis function, and $\xi_i$ is the coefficient of $i$-th control point. The knot vector is 
$\{t_0,t_1,t_2,\dots, t_{k+n} \}$.  $B_{i,k} (t)$ is calculated recursively by the following formula:
\begin{align}
    \label{basis}
    B_{i, 0} & = \left \{
    \begin{matrix}
           1, & t_i \leq t < t_{i+1}, \\
           0, & \text{otherwise},
    \end{matrix}
    \right.\nonumber \\
    B_{i,k} &= \frac{t-t_{i}}{t_{i+k}-t_i} B_{i, k-1}(t) + \frac{t_{i+k+1}-t}{t_{i+k+1}-t_{i+1}}B_{i+1,k-1}(t).
\end{align}
In this study, we apply the B-spline to parameterize the time-varying parameters in the GODE. The degree of the basis function, $k$, controls the complexity of the time-varying effect; the number of the control points, $n$, controls the number of parameters in the GODE. If $k=0,n=1$, equation (\ref{bspline}) can be rewritten as
\begin{equation}
    \label{basis1}
    \beta(t)=B_{0,0}(t)\xi_0,
\end{equation}
where 
\begin{equation}
    \label{basis2}
    B_{0,0}(t) = \left \{
    \begin{matrix}
    1, & t_0 \leq t < t_{1}, \\
    0, & \text{otherwise}.
    \end{matrix}\right.
\end{equation}
In this case, the knot vector is $\{t_0,t_1 \}$, i.e., the first and endpoints of the integration time interval. And, equation (\ref{basis1}) and (\ref{basis2}) are equivalent to $\beta(t) = \xi_0$, and $B_{0,0} (t)=1$. Therefore, the Neural-GODE model is reduced to the standard Neural-ODE model with constant parameters when $k=0, n=1$. On the other hand, increasing the number of control points (knots), $n$,  increases number of parameters in the integration interval. If an ODE is solved by the simple Euler method with small step size, the increased number of knots makes each Euler step have different parameters, i.e., the increased number of different  $\theta_t$ in equation (\ref{restrans}). Since the ResNets can be interpreted as nonlinear ODE as equation (\ref{restrans}), the Neural-GODE with time-varying parameters can represent the ResNets in this case. 

\subsection{Model architectures}\label{subsec22}
We experiment with the residual networks (ResNets) architecture that has been used for comparison in Chen et al. 2018 \cite{chen2018neural}, see Fig. 1. The ResNets first down-samples the input three times with convolution layers, then applies multiple standard residual blocks \cite{he2016deep}, see Fig. 1 (left panel). Each residual block consists of two convolution layers with a kernel size of  3. The standard neural-ODE model and neural-GODE model replace the residual blocks with an ODE block, see Fig. 1 (right panel). However, the ODE block of the Neural-GODE is a system with time-varying parameters, which makes the model more 
flexible for training. 

\begin{figure}[h]%
\centering
\includegraphics[width=0.5\textwidth]{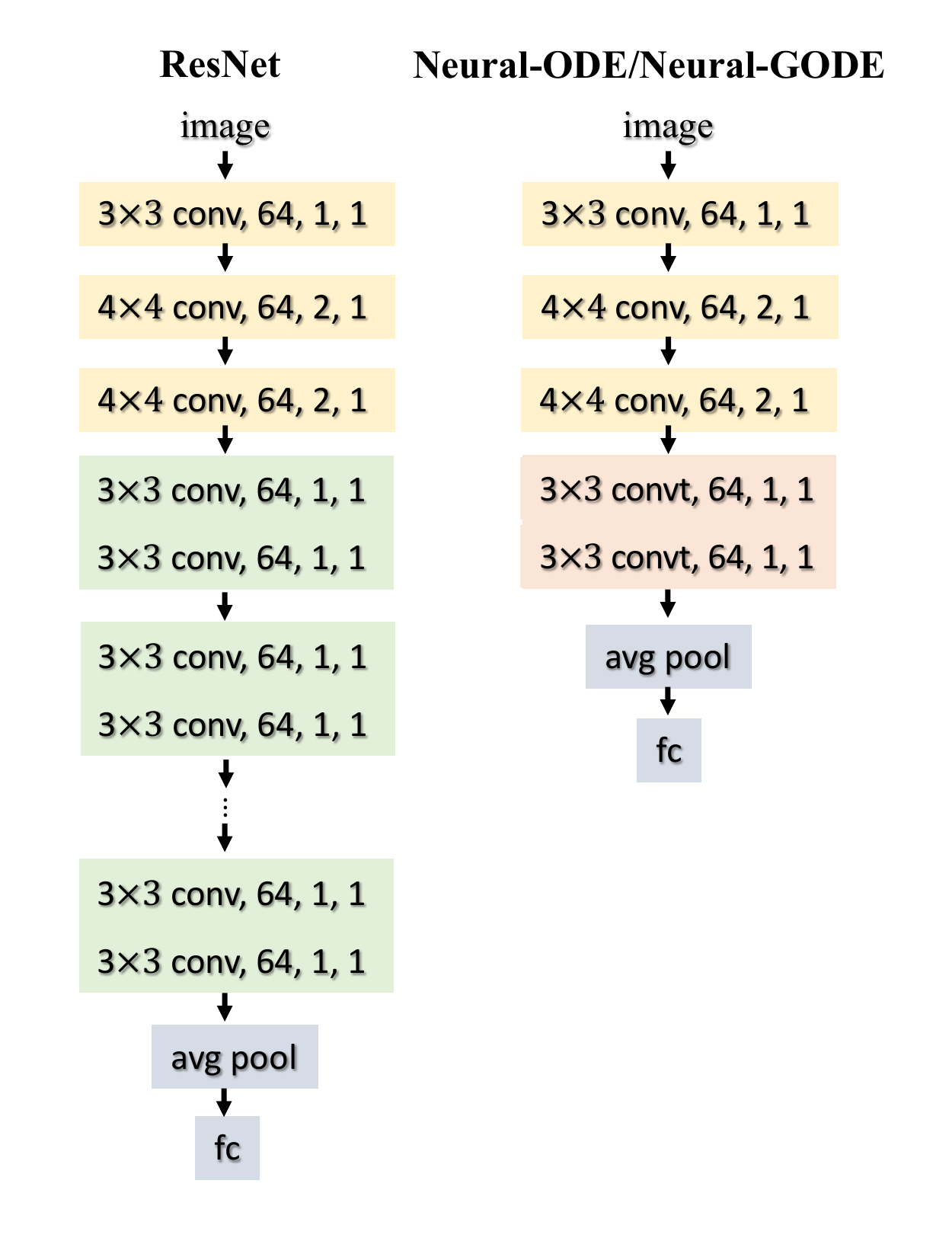}
\caption{Model architectures.}\label{fig1}
\end{figure}

\subsection{Implementation}\label{subsec23}
The implementation of the ResNets, Neural-ODE, and Neural-GODE models follow the MNIST training in \cite{chen2018neural}. The training images are transformed by randomly cropping with padding of 4 on each border. We apply the group normalization over each mini-batch \cite{wu2018group} right after each convolution layer and before the activation. We initialize the weights of the B-spline parameterized filter in the customized convolution layer (the module “convt” in Fig. 1). We use SGD with a mini-batch size of 128 in training. The learning rate starts from 0.1 and is divided by 10 at epochs 60, 100, and 140. A total of 160 epochs are trained for each model. We use momentum of 0.9. The dropout is not applied, following the practice in \cite{ioffe2015batch}. We apply the Euler method with a step size of 0.05 for solving the ODE systems both in the Neural-ODEs and Neural-GODE by using package torchdiffeq in \cite{paszke2017automatic}. 
In testing, random cropping is not applied, and a mini-batch size of 1000 is used. The model performance is reported based on the testing dataset using accuracy as the metrics. All the experiments are implemented on GPU (Tesla V100 with 16GB G-Memory), programming code can be found at \url{https://github.com/Duo-Yu/Neural-GODE}. 

\section{Experimental Results}\label{sec3}
We evaluate the Neural-GODE on two benchmark datasets, i.e. MNIST and CIFAR-10. 

\subsection{Model performances}\label{subsec31}

The MNIST and CIFAR-10 are standard benchmark datasets for computer vision and deep learning. In the MNIST, images are white-black handwritten digits with a size of 28×28 pixels. The MNIST classification aims to predict the ten handwritten digits. It has 70, 000 images with 60, 000 samples in the training dataset and 10, 000 in the testing dataset. The CIFAR-10 dataset is 60, 000 colored images with 10 classes. Each image has 32×32 pixels. Generally, the training set of CIFAR-10 consists of 50, 000 samples. The ResNets is implemented with 6 and 20 residual blocks on MINIST and CIFAR-10, respectively. With such numbers of residual blocks can reach more than 99\% accuracy in training on both datasets. The ODE systems are solved from 0 to 1 in the standard Neural-ODE and the proposed Neural-GODE. The number of control points of B-spline in Neural-GODE, n, is 4 and 8, respectively, on MNIST and CIFAR-10 training. The degree of the B-spline is 1. The effect of these hyperparameters, including the B-spline order, degree, and the integration interval of ODE systems, is shown in the following subsection, Table II. 
For test error, the Neural-GODE has the best performance, see Table I. The ResNet has a slightly lower accuracy than that of Neural-GODE. The standard Neural-ODE has the worst performance compared to the other two models. Especially in the classification task based on CIFAR-10, which is more complex than the MNIST. The standard Neural-ODE model has about 2\% lower accuracy than the Neural-GODE, Table I. In terms of training efficiency, although the standard Neural-ODE and Neural-GODE are slower than ResNets, they show higher memory efficiency. Both Neural-ODE and Neural-GODE consist of a smaller number of training parameters than that of ResNets, see Table I. In summary, the  Neural-GODE model has the advantage in both predictive accuracy and memory efficiency compared to the other two models.

\begin{table*}[h]
\begin{center}
\begin{minipage}{\textwidth}
\caption{Model performance comparison}\label{tab1}%
\begin{tabular}{@{}lllll@{}}
\toprule
\multicolumn{2}{c}{Model} & Neural-GODE  & ResNets & Neural-ODE\\
\midrule
\multirow{3}{*}{MNIST} 
& Test error (\%) & 0.31& 0.33 & 0.40\\
& \# params (M) & 0.43 & 0.57 & 0.21 \\
& Time/iteration (s) &0.035 & 0.012 & 0.038\\ 
\midrule
\multirow{3}{*}{CIFAR-10} 
& Test error (\%) & 13.49 & 13.47& 15.32\\
& \# params (M) & 0.72  & 1.6  & 0.21  \\
& Time/iteration (s) &0.038 & 0.026& 0.041\\ 
\botrule
\end{tabular}
\end{minipage}
\end{center}
\end{table*}

\begin{table}[h]
\begin{center}
\begin{minipage}{174pt}
\caption{Effect of time-varying parameters}\label{tab2}%
\begin{tabular}{@{}llllll@{}}
\toprule
n & k & T & \#params & Time(s) & Test error (\%)\\
\midrule
2& 1& 1 & 281,738 &	0.039	&14.68\\
4& 	1& 	1& 	429,194	& 0.038& 	14.19\\
6& 	1& 	1& 	576,650& 	0.037& 	13.97\\
8& 	1& 	1& 	724,106	& 0.038& 	13.49\\
10& 	1& 	1& 	871,562	& 0.037& 	13.65\\
12& 	1& 	1& 	1,019,018& 	0.038	& 13.64\\
\midrule
8	&2	&1	&724,106	&0.038	&14.16\\
8&	3	&1	&724,106&	0.036&	14.10\\
8&	4	&1	&724,106&	0.036	&13.80\\
8&	5	&1	&724,106&	0.037	&14.22\\
\midrule
8&	1&	2&	724,106	&0.064	&13.79\\
8&	1&	3&	724,106&	0.093&	13.84\\
\botrule
\end{tabular}
\end{minipage}
\end{center}
\end{table}
The main difference between Neural-GODE and Neural-ODE is that the ‘convt’ layer weightes are parameterized with B-spline functions of integration time (t) in Neural-GODE rather than constant in the Neural-ODE. As an illustration, we plot the weights of the first convolution layer of the residual block in the ResNets, the first ‘convt’ layer of the ODE blocks in the Neural-GODE and Neural-ODE in Fig. 2. We observe that the kernel weights from the Neural-ODE are constant across the integration time (red lines in Fig. 2) as expected. The weights in the Neural-GODE (blue lines in Fig. 2) and ResNets (green lines in Fig. 2) vary across layers or integration time, and the weights of the Neural-GODE are smoother than those of the ResNets, which might be the reason why the proposed Neural-GODE model could outperform the ResNets and Neural-ODE models in terms of computing efficiency and prediction accuracy. 

\begin{figure*}[h]%
\centering
\includegraphics[width=0.9\textwidth]{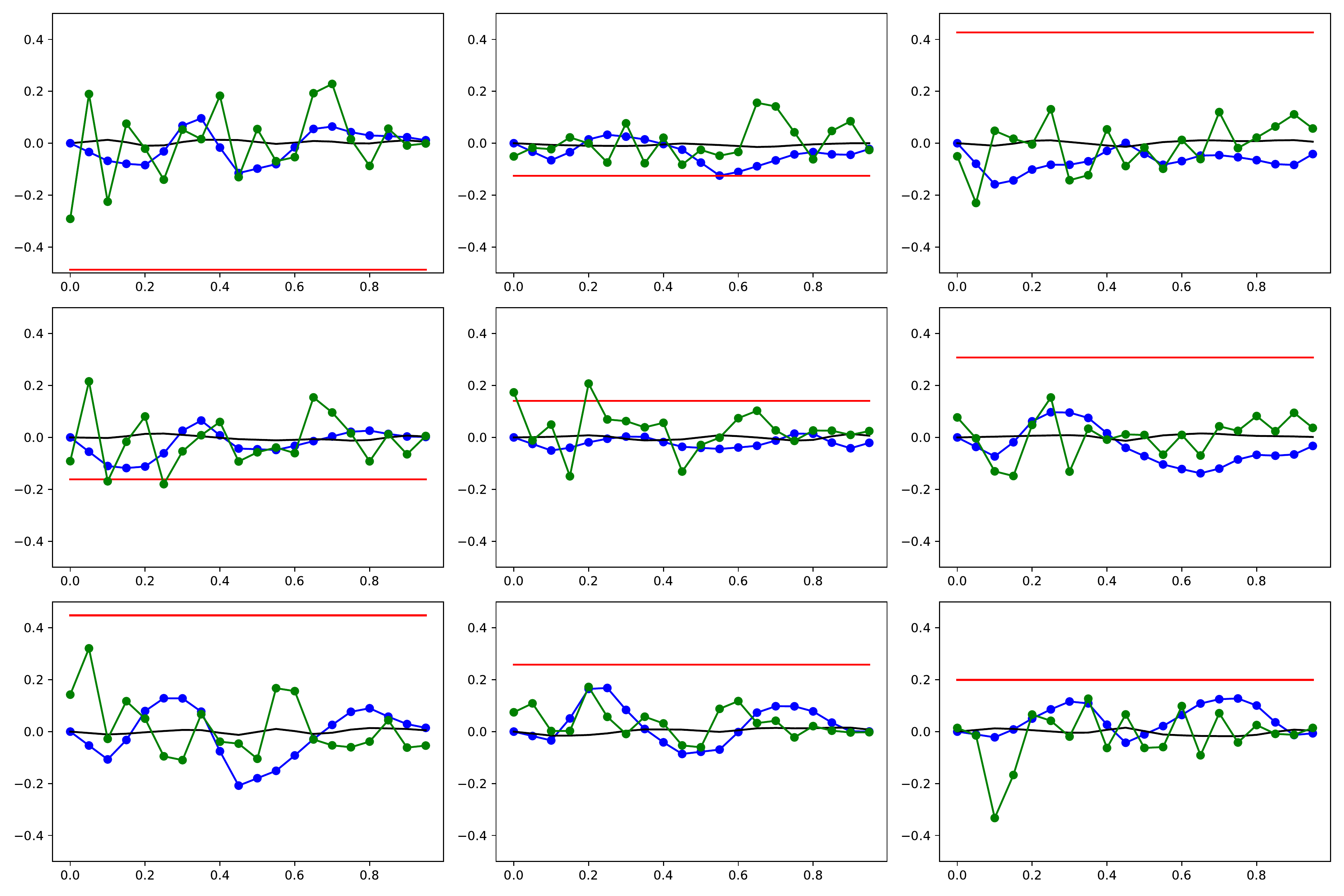}
\caption{The patterns of selected estimated weights of a 3×3 kernel from three models from the CIFAR-10 data example. The x-axis represents the index of layers in ResNets, and the integration time in Neural-GODE and Neural-ODE. The red line denotes the weight of the first ‘convt’ layer in the ODE block of the Neural-ODE; the blue line is the weight of  the first ‘convt’ layer in the ODE block of the Neural-GODE; and the green line is the weight of the first convolution layer in the residual blocks of the ResNets.}\label{fig2}
\end{figure*}

\begin{table*}[h]
\begin{center}
\begin{minipage}{\textwidth}
\caption{ODE solver comparison}\label{tab3}%
\begin{tabular}{@{}llllll@{}}
\toprule
\multicolumn{2}{c}{Model} & \multicolumn{2}{c}{Neural-GODE} & \multicolumn{2}{c}{Neural-ODE} \\
\midrule
& & Dopri5& Euler & Dopri5 & Euler \\
\multirow{2}{*}{MNIST} 
& Test error (\%) & 0.37 & 0.31 & 0.37 & 0.40\\
& Time (s) &0.153&	0.035	& 0.063 & 0.038\\
\midrule
\multirow{2}{*}{CIFAR-10}
& Test error (\%) & 13.99&	13.49& 15.40 &	15.32\\
& Time (s) &0.224&	0.038	&0.072&	0.041 \\
\botrule
\end{tabular}
\end{minipage}
\end{center}
\end{table*}

\subsection{Layer-varying parameters }\label{subsec32}

To evaluate the effect of hyperparameters of Neural-GODE on prediction results, such as the number of control points ($n$) and degree ($k$) of B-spline as well as the integration interval endpoint ($T$) of ODE, we use different settings to train the Neural-GODE on CIFAR-10 data. We find that there exist an optimal number of control points for B-spline, which directly affects the Neural-GODE model size (the number of parameters). Based on CIFAR-10, the optimal number of control points is 8 when the ODE is solved by the Euler method with a step size of 0.05, see Table II. If we increase the number of control points from 2 to 8, the test error is reduced by 1\%. However, if the number of control points increases from 8 to 12, the test error increases slightly. According to the experiment, the linear B-spline fits the data well, i.e., when the degree of B-spline ($k$) is 1, the prediction accuracy reaches the highest. If we increase the endpoint ($T$), the training speed will significantly decrease. For example, given the linear B-spline with 8 control points, if the end integration time increases from 1 to 3, each iteration training time increases from 0.038s to 0.093s, which is about 2.4 times increase, see Table II. At the same time, we do not find much benefit in terms of prediction accuracy when we increase the endpoint of ODE. 

\subsection{ODE solvers}\label{subsec33}
In the experiment, we mainly focused on the Euler method to solve the ODEs because of its first-order equivalence to the general residual network (see Method section). However, we could use alternative ODE solvers in the neural ODE models. For example, instead of using the fixed-step method such as Euler, we also implement another adaptive-step method, i.e., Runge-Kutta of order 5 of Dormand-Prince-Shampine (dopri5 in the package \textit{torchidffeq}). We observe that the Runge-Kutta and Euler methods perform similarly in predictive accuracy (see Table III). For the MNIST data set, both Neural-ODE and  Neural-GODE produced 0.37\% of the test error when the Runge-Kutta method was applied. Using the Euler method with a step size of 0.05, the Neural-GODE has a slightly lower test error than the standard Neural-ODE. However, the Neural-GODE model shows lower test errors than the standard Neural-ODE using either the Runge-Kutta or the Euler methods. In terms of training efficiency, the Runge-Kutta method is significantly slower than the Euler method, which is the main reason that we use the Euler method as the ODE solver in our experiment, see Table III. Similar comparison results are observed from training experiments based on CIFAR-10. 

\subsection{ODE function forms}\label{subsec34}

In this section, we investigate the effect of the number of customized CNN layers inside the ODE block of Neural-GODE. The standard Neural-ODE directly replaces the residual blocks with one ODE block consisting of two convolution layers. The "time" t  is considered as a separate channel and concatenated with other image channels. The concatenated channels are the input of the ODE block in which the convolution functions are standard. Instead of combining the "time" t with the convolution layer input, we parameterize t inside the convolution function. Specifically, time-varying kernels are used rather than using the kernels with constant weights. Besides the number of control points of the B-spline, the number of customized CNN layers can also affect the total number of parameters and the complexity of the ODE block. We compare the predictive performance given the different number of control points and the number of customized CNN layers based on CIFAR-10. The fixed parameters include the degree of B-spline (k = 1), the integration interval ([0, 1]), and the ODE solver (Euler method with a step size of 0.05). We observe a trade-off between the number of control points and the number of customized CNN layers of the ODE block. When the number of control points of the B-spline is small, a larger number of CNN layers may increase the predictive performance. For example, if the number of control points is 2 or 4, better predictive accuracy can be reached when the number of CNN layers is 3, Table 4. However, if we further increase the number of control points, two layers of CNN have better performance. Since the ODE block with more CNN  layers takes a longer time to solve, we apply two layers of CNN with eight control points of the B-spline for CIFAR-10 training.

\begin{table}[h]
\begin{center}
\begin{minipage}{174pt}
\caption{Effect of the number of the customized CNN layers.}\label{tab4}%
\begin{tabular}{@{}lllll@{}}
\toprule
n & \# layers & \#params & Time (s) & Test error(\%)\\
\midrule
\multirow{4}{*}{2} 
&1&	207,754	&0.024&	15.83\\
&2&	281,738	&0.039&	14.68\\
&3&	355,594	&0.047&	14.31\\
&4&	429,450	&0.061	&14.37\\
\midrule
\multirow{4}{*}{4} 
&1	&281,482&	0.024&	16.26\\
&2&	429,194	&0.038&	14.19\\
&3&	576,778	&0.047&	13.89\\
&4&	724,362	&0.062&	14.40\\
\midrule
\multirow{4}{*}{6}
&1&	355,210	&0.025&	15.67\\
&2&	576,650	&0.037&	13.97\\
&3&	797,962	&0.048&	14.07\\
&4&	1,019,274&	0.062	&14.59\\
\midrule
\multirow{4}{*}{8}
&1&428,938&	0.025	&15.41\\
&2&	724,106	&0.038&	13.49\\
&3&	1,019,146&	0.047&	13.95\\
&4&	1,314,186&	0.063&	14.02\\
\botrule
\end{tabular}
\end{minipage}
\end{center}
\end{table}

\section{Conclusion}\label{sec4}

The idea of bridging deep residual networks (ResNets) with the discretized ordinary differential equations (ODE) has raised many interests in the deep learning research field recently \cite{haber2017stable, chang2018reversible, lu2018beyond, li2017maximum, li2019deep,chen2018neural}. With the well-established ODE properties, theories, and numerical solutions, novel deep learning architectures and training algorithms have been proposed based on such connection. In this study, we further explore the performance of Neural-ODE based on two benchmark classification tasks, i.e., the classification problems of MNIST and CIFAR-10. We confirm that the Neural-ODE model has a training efficiency advantage but compared to the deep ResNets in terms of training memory as stated in Chen et al. \cite{chen2018neural}. 
On the other hand, the predictive accuracy is not as good as the ResNets. For the CIFAR-10 dataset, the test error of the standard Neural-ODE is 15.32\% with 0.21 million parameters, while the test error of the ResNets is 13.73\% with 1.6 million parameters, see Table I.

To overcome the predictive accuracy disadvantage of the standard Neural-ODE, we propose a time-varying Neural-GODE, which improves the model flexibility and eventually improves the predictive performance. Instead of using constant weights of CNN layers from the standard Neural-ODE, we parameterize the kernel weights with time-varying parameters. Specifically, the weights are B-spline functions of "time" t. As a result, the proposed Neural-GODE model reaches similar or slightly better prediction accuracy than ResNets.  

The ResNets model is initially developed for image recognition [1]. For the CIFAR-10 data analysis, He et al. designed a sophisticated deep ResNets model which can eventually reach a test error of 6.43\% with the 110 layers, and the first layer was 3×3 convolutions. Then they used a stack of 6n layers with 3×3 convolutions on the feature maps of sizes {32,16,8}, respectively, with 2n layers for each feature map size, where n=18. In our new Neural-GODE model, the input image is directly down-sampled by three layers of 3×3 convolutions, then processed by one GODE block. The map size in the ODE block is 8. To reach a lower test error using the Neural-GODE model, we will explore multiple ODE blocks with different map sizes in future research.  

With the benefits of training efficiency and accuracy, the proposed Neural-GODE model can be applied in various tasks. It is straightforward to apply the Neural-GODE in any image recognition problem, such as the radiology images in electronic health records (EHR). On the other hand, the Neural-GODE can be applied in time-series analysis for one-dimensional data inputs \cite{bonnaffe2021neural}, which will also be our future applied research. 

In statistical literature, the neural networks (deep learning models) are considered as the non-parametric regression models. As early as the year that a neural network with one hidden layer was proved to have the universal approximation property \cite{hornik1989multilayer}, White has evaluated the asymptotic results of back-propagation learning in single hidden-hidden layer feedforward network models \cite{white1989some}. Also, Shen et al. studied the asymptotic properties, including the consistency, convergence rate, and asymptotic normality for the neural network sieve estimators with one hidden layer \cite{shen2019asymptotic}. Recently, theoretical properties of deep learning models have been explored from a statistical perspective ([48-50]. However, the theoretical properties have not been established for more complicated deep learning models, such as recurrent neural networks (RNN), convolutional neural networks (CNNs) (including ResNets), and Neural-ODE. We have established some asymptotic results for the sieve estimators of the ODE with both constant and time-varying parameters \cite{xue2010sieve}. The large-sample properties for the GODE model have also been investigated in our previous work \cite{miao2014generalized}. These theoretical results have laid a solid foundation to establish the asymptotic properties for the proposed Neural-GODE model, in which future research is warranted.   

\section*{Acknowledgements}
This work was supported in part by NIH grant R01 AI087135(HW),grant from Cancer Prevention and Research Institute of Texas (PR170668) (HW), grant NSF/ECCS 2133106 (HM), and NSF/DMS 1620957 (HM). 

\bibliography{neuralGODE}

\end{document}